\documentclass[conference]{IEEEtran}
\IEEEoverridecommandlockouts
\usepackage{cite}
\usepackage{amsmath,amssymb,amsfonts}
\usepackage{algorithmic}
\usepackage{graphicx}
\usepackage{textcomp}
\usepackage{xcolor}
\usepackage{multirow}
\usepackage{caption}
\usepackage{subcaption}
\def\BibTeX{{\rm B\kern-.05em{\sc i\kern-.025em b}\kern-.08em
    T\kern-.1667em\lower.7ex\hbox{E}\kern-.125emX}}

\usepackage{fancyhdr}
\begin{document}

\title{A New Perspective on Smiling and Laughter Detection: Intensity Levels Matter}
\author{\IEEEauthorblockN{Hugo Bohy\textsuperscript{\textsection}}
\IEEEauthorblockA{\textit{ISIA Lab} \\
\textit{University of Mons}\\
Mons, Belgium \\
hugo.bohy@umons.ac.be}
\and
\IEEEauthorblockN{Kevin El Haddad\textsuperscript{\textsection}}
\IEEEauthorblockA{\textit{ISIA Lab} \\
\textit{University of Mons}\\
Mons, Belgium \\
kevin.elhaddad@umons.ac.be}
\and
\IEEEauthorblockN{Thierry Dutoit}
\IEEEauthorblockA{\textit{ISIA Lab} \\
\textit{University of Mons}\\
Mons, Belgium \\
thierry.dutoit@umons.ac.be}
}
\newcommand{\snl}{S\&L }
\maketitle

\begingroup\renewcommand\thefootnote{\textsection}
\footnotetext{Equal contribution}
\endgroup

\thispagestyle{fancy}

\begin{abstract}

Smiles and laughs detection systems have attracted a lot of attention in the past decade contributing to the improvement of human-agent interaction systems. But very few considered these expressions as distinct, although no prior work clearly proves them to belong to the same category or not. In this work, we present a deep learning-based multimodal smile and laugh classification system, considering them as two different entities. We compare the use of audio and vision-based models as well as a fusion approach. We show that, as expected, the fusion leads to a better generalization on unseen data. We also present an in-depth analysis of the behavior of these models on the smiles and laughs intensity levels. The analyses on the intensity levels show that the relationship between smiles and laughs might not be as simple as a binary one or even grouping them in a single category, and so, a more complex approach should be taken when dealing with them. We also tackle the problem of limited resources by showing that transfer learning allows the models to improve the detection of confusing intensity levels.

\end{abstract}


\section{Introduction}
With the growth of virtual agents and other human-centric applications, detection systems for nonverbal expressions have been attracted the attention of the research community in the past decade, especially concerning smiles and laughter~(S\&L). This is due to the importance of these expressions in human communications.
Indeed they not only have emotional but also social functionalities: according to \cite{glenn2003laughter}, \snl tend to happen during interactions with others rather than alone. They can control the flow of a conversations: change the current topic\cite{holt2010last, provine1993laughter} or encourage a person to carry on speaking\cite{hayward2018smile}. Laughter can be contagious to listeners and can lighten the mood of the conversation\cite{scott2014social}. \snl are also expressions used frequently in human-human interactions. Indeed the ICSI corpus\cite{janin2003icsi} counts about 10\% of its total verbalizing time as being laughter \cite{gilmartin2013exploring, laskowski2007analysis} and Chovil in \cite{chovil1991discourse} reports not even considering smiles in the study due to its high frequency of occurrence in the data compared to other expressions.

It is therefore not surprising that the detection of laughs or smiles became an attractive field rising alongside the deep learning technologies and AI-backed human-agent interaction systems.

A plethora of work can be found on smile detection. We estimate that the vast majority of them are based on the visual cue as we could find very few work based on other modalities \cite{guo2018smile, whitehill@smile, cui2018elm}, notably the audio cue was rather absent from the state-of-the-art although smiles were proven to be recognizable audible \cite{tartter1994hearing, arias2018hearing, arias2018@hearing2}

Fewer work can be found on laughter detection. They focus on the audio and the visual modalities individually but also in multimodal approaches. Kantharaju et. al. in \cite{kantharaju2018automatic} present an automatic detection of different categories of laughter using audio-visual data. The authors in \cite{7298420} use full-body motion capture data to detect laughter while \cite{berker@laugh} investigates the laughter detection based on audio and facial motion capture.

Surprisingly, very few work can be found where \snl are considered as two distinct expressions, and none of them attempts to classify/detect them as different entities. Indeed, even though the authors in \cite{devillers@snl} annotate them as two expressions in their work, they build classifiers considering them as the same class. The authors in~\cite{ito@snl} propose a system classifying smiles vs non-smiles based on the visual cue and laugh/non-laugh based on a single modality and on multimodal data, but no smile/laugh discrimination is presented. One reason for this might be the difficulty for the models to learn the differences between smiles and laughs, especially given the limited amount of resources available. Another reason might be the common representation for some, of smiling being a less intense expression of laughter or both even both being the same expression, which is to the best of our knowledge, unproven yet.

Although \snl are commonly defined as the former being a purely facial expression while the latter being an audio-visual one, no clear answer can be found in the literature as of the relationship between these two: are they the same expression at different intensity levels~? Are they distinct expressions although smiles can be perceived in laughs~? Or does it depend on the context/situation~? Ruch and Ekman observe in \cite{ruch2001expressive} that enjoyment smiles were involved in laughter while Trouvain's perception study in \cite{trouvain2001phonetic} revealed that some participants preferred to categorize speech-laughs into smiles and laughs (speech-laughs being laughter-speech co-articulation phenomenon involving laughter intermingling with speech). In \cite{davilaross@snlrelation} the authors present existing relationships between \snl on several levels, suggesting a common ancestry of these expressions and therefore that at least some relationship exist between them.

Since no study showing proof of smiles and laughs being separated expressions exists, nor a smiling-laughter continuum established, we consider it important, in this work, to approach \snl as two distinct expressions. Doing this makes it easier to analyse the common and differentiating points, and allows us to leverage the feature extraction power of deep learning to further examine the intrinsic problems of smiles and laughter detection systems.  

In this paper, we present several contributions. First, we propose a first step towards an efficient \snl detection system discriminating between laughs and smiles, as opposed to systems detecting a single category of smiles and laughs, by building and analysing classifiers based on the audio and the visual cues. Given the observations mentioned above regarding the relationship between \snl, we push the analyses further by examining the behavior of models with regard to \snl intensity levels, while being trained without any supervised knowledge of these intensities. These analyses reveal that not all \snl levels are equal in the eyes of deep learning systems. This understanding might change the simplistic approaches taken for building laughter or smiling detection systems.

\snl are difficult to collect in naturalistic setups and to annotate. This difficulty to access accurately annotated data represents a challenge to develop efficient systems and a significant barrier of entry for new contributions in the field. Indeed this lack of well annotated data makes it more difficult to leverage the efficiency of deep learning methods, and thus stalls the improvement of \snl detection systems.
So, as a second main contribution, we apply transfer-learning by leveraging the knowledge learned with speech data by the models to improve the efficiency and generalisation capabilities of the models for \snl detection. 

The following is a more detailed summary of our main contributions in this work:
\begin{enumerate}
    \item we propose the first deep learning-based \snl classification system that we know of that considers \snl as two different entities
     \item a deeper analysis than what can be found in the literature of the model's behavior showing that deep learning-based systems implicitly take into account the \snl intensity levels in their learning process without being trained with any explicit knowledge of them
     \item we show that transferring knowledge from visual lipreading task and from audio word classification improves the performance of the models and help tackle the problem of limited resources
\end{enumerate}

The paper is organised as follows: in Section~\ref{dataset} we present the datasets used for our experiments. Section~\ref{classifier} contains the description of the model architectures for the audio and the visual modalities, as well as for the fusion of both. We describe the experimental protocol followed to train the models in Section~\ref{experiment} and we discuss the results of the aforementioned experiments in Section~\ref{results}.

\section{Dataset}
\label{dataset}
The data used here are subsets of the Nonverbal Dyadic Conversation on Moral Emotions~(NDC-ME) \cite{heron2018dyadic}, and of the IFA Corpus~(IFADV) \cite{van2001ifa} for which the \snl were annotated.
The \snl were segmented and the intensity level was added to each segment. We followed the annotation protocol described in \cite{elhaddad@snldynamics}. 
The annotations were made using the ELAN software \cite{elan} by two annotators on average and are available to the community~\cite{elhaddad@ib}.

NDC-ME is an audiovisual collection of dyadic interactions focusing on the emotions expressed during speaker-listener interactions. The subset we use is distributed in 17 dyadic interactions split between 10 male and 4 female individuals, with 7 male-male, 6 male-female and 4 female-female pairs. During these interactions, each duo discusses emotional topics introduced by an open question. Since some of those interactions are not fully annotated, the total duration of annotated data is about 90 minutes with an unbalanced distribution between individuals.

IFADV is also a collection of audio-visual recordings of dyadic conversations. The subset we used contains 23 dyadic interactions of 15 male and 28 female individuals with 4 male-male, 8 male-female and 11 female-female pairs of interactions. The annotations cover only the first two minutes of each file, leading two around 46 minutes of annotated data.

The laughs intensities are divided in three levels (low, medium and high) and the smiles intensities in four (subtle, low, medium and high). According to the authors of the previously mentioned papers, the subtle level was added to capture all the levels of smiles even the ones that are normally left out because of the difficulty to annotate them: subtle smiles co-occurring with other expressions for instance. A third class, referred to as the \textbf{None} class, includes all segments of the recordings that contain neither laughter nor smiles, such as neutral expressions and speech. Therefore we ended up with three main classes \textbf{Laughs}, \textbf{Smiles} and \textbf{None}, which will be used for training without taking into account the intensity levels.

\begin{figure*}[htbp]
    \centering
    \includegraphics[width=0.9\textwidth]{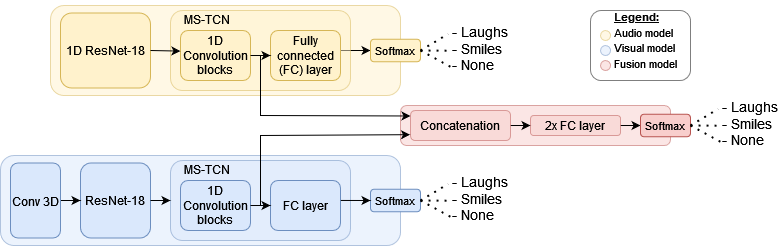}
    \caption{Architectures of the audio model, the visual model and the fusion of both.}
    \label{fig:architectures}
\end{figure*}

\section{Classifiers/Detection systems}
\label{classifier}
In this section, we describe the deep learning architectures used for classification. Since \snl are distinguishable through both audio and visual, we first separated our system in two models, one per modality. We used the audio and visual models proposed in \cite{martinez2020lipreading} and \cite{ma2020towards}, which present audio word recognition/classification and visual lip reading applications. We then perform the fusion of both modalities. Fig.~\ref{fig:architectures} displays a schematic representation of the system architecture. 

There were several intuitions behind these choices. The first one is the fact that modalities, as mentioned before, could be a major factor helping a model discriminate between smiles, laughs and everything else. This type of architecture already takes this aspect into account and successfully applies it on speech, making it a good candidate for our experiments. The second intuition is the fact that our classification learning could potentially benefit from knowledge learned from speech. Indeed by learning to recognise speech pattern, these models learn audio and visual features as well as pattern that could also be useful for classifying smiling, laughter and others. In fact, given the limited amount of \snl data available, a transfer learning technique is a good option to optimise learning with a small dataset. Transfer learning based on facial recognition has already been used successfully in the context of smiles detection \cite{xin@smile}.

\subsection{Audio Cue}
Our audio model is a mix between the backbone architecture proposed in \cite{martinez2020lipreading} for speech audio data and the frontend layers proposed in \cite{ma2020towards}. The model consists of a modified 18-layer ResNet backbone using 1D kernels fed to a multi-scale temporal convolutional network (MS-TCN). We used 1D operations since audio waveforms are unidimensional signals. The MS-TCN is a multi-layered combination of 1D convolutions with batch normalisation and PReLU activation layers, designed in such a way that it models both short and long term temporal information simultaneously. The final layer of the MS-TCN block is a fully connected (FC) layer with a softmax activation function to perform the classification.

\subsection{Visual Cue}
We used the architecture proposed in \cite{ma2020towards} for the visual recordings. The standard ResNet18 backbone was modified to have its first layer to be a 3D convolution of kernel size $5\times7\times7$. This layer extracts spatiotemporal features from the sequences of image given as input to the model. The backbone output is fed to the MS-TCN block of layers described above. In the same fashion as for the audio, we used a softmax activation function after the MS-TCN block to classify data as \textbf{Laughs}, \textbf{Smiles} or \textbf{None}.

\subsection{Fusion}
We performed the fusion by feeding the output of each modality to a network of two fully connected layers of size 1024 and 3 respectively. We froze the weights of each modality and we used the concatenation of their output as input of the fusion models.

\section{Experiments}
\label{experiment}
For our experiments, we extracted from the NDC-ME data 8,352 videos with 1.22s windows overlapping by 0.4s and split them with respect to their classes giving us 446 \textbf{Laughs},  4,858 \textbf{Smiles} and 3,048 \textbf{None}. We re-sampled the audio data at 16 kHz and we converted the visuals from RGB to grey scale and then extracted a 96 x 96 pixel region of interest~(ROI) around the mouth. We then distributed 70\% of each class as training data, 15\% as validation data and 15\% as test data. Classes were balanced during partitioning using a random weighted batch sampler with each class given a weight proportional to the inverse of the number of elements in the class. 

We conducted three different training sessions for each modality using exclusively the NDC-ME subset: a training with weights randomly initialised (referred to as \textit{from scratch} models), an other one where we trained all layers from the pre-trained lipreading model (\textit{fully fine-tuned} models) and the last one where we trained only the MS-TCN related layers of the pre-trained model (\textit{last layers fine-tuned} models). Each training session consisted of 80 epochs with an evaluation of the model at the end of each epoch using the validation partition. Since we had limited amount of data, we used an initial learning rate of $3 \times 10^{-6}$ and a batch size of 16. The learning rate decreased using a cosine annealing schedule. The models were then evaluated with the test partition of our NDC-ME subset on the one hand and on the other hand with all annotated IFADV data to assess the generalisation of the models.

For the fusion, we trained our network of fully connected layers initialised with random weights using several configurations of audio and visual models. In this paper, we present only two configurations: the first one combines both modalities trained from scratch while the second configuration is based on the results obtained for each modality individually. This second fusion model combines the output from the audio model fine-tuned on its MS-TCN layers only and the visual model fine-tuned on all layers. We used an initial learning rate of $3 \times 10^{-6}$ and a batch size of 16, and the training lasted 30 epochs.
The other possible combinations were indeed tested during the experimentation phase, but did not bear any interesting results (either performed generally less well than the ones presented here or from which we could not draw any relevant conclusion).

\section{Results and Discussions}
\label{results}
\subsection{Results}

Table~\ref{tab:PRFU} contains the Precision, Recall, F1-score and UAR metrics for the S\&L classifications. The configurations are separated by modality and evaluation dataset. The highest value per metric and per separation is highlighted in green. We can observe better generalisation to other datasets for the fine-tuned models compared to the models trained from scratch. Fine-tuning is also better for classification except on the NDC-ME evaluation of the fusion model.

\begin{table*}[htbp]
\centering
\begin{center}
\caption{Precision, Recall, F1-score and UAR metrics per configuration for S\&L classification. The configuration name follows a XYZZZ pattern with X being the modality: A (Audio), V (Video) or F (Fusion); Y the training method: S (from Scratch) or F (corresponding Finetuning method); ZZZ the evaluation dataset: NDC (NDC-ME) or IFA (IFADV). The highest value for each configuration and dataset is coloured in green.}
\label{tab:PRFU}
\begin{tabular}{|l|llll|llll|llll|}
\hline
\multicolumn{1}{|c|}{}                                   & \multicolumn{4}{c|}{\textbf{AUDIO}}                                                                                                                                  & \multicolumn{4}{c|}{\textbf{VIDEO}}                                                                                                                            & \multicolumn{4}{c|}{\textbf{FUSION}}                                                                                                                  \\ \cline{2-13} 
\multicolumn{1}{|c|}{\multirow{-2}{*}{\textbf{Metrics}}} & \multicolumn{1}{c|}{\textit{ASNDC}} & \multicolumn{1}{c|}{\textit{AFNDC}}                & \multicolumn{1}{c|}{\textit{ASIFA}} & \multicolumn{1}{c|}{\textit{AFIFA}} & \multicolumn{1}{c|}{\textit{VSNDC}} & \multicolumn{1}{l|}{\textit{VFNDC}}                & \multicolumn{1}{l|}{\textit{VSIFA}} & \textit{VFIFA}                & \multicolumn{1}{c|}{\textit{FSNDC}} & \multicolumn{1}{c|}{\textit{FFNDC}} & \multicolumn{1}{c|}{\textit{FSIFA}} & \multicolumn{1}{c|}{\textit{FFIFA}} \\ \hline
Precsion                                                 & 0.4828                              & \multicolumn{1}{l|}{{\color[HTML]{009901} 0.4937}} & 0.3431                              & {\color[HTML]{009901} 0.3719}       & {\color[HTML]{009901} 0.7116}       & \multicolumn{1}{l|}{0.7074}                        & 0.3993                              & {\color[HTML]{009901} 0.4137} & {\color[HTML]{009901} 0.6154}       & \multicolumn{1}{l|}{0.6018}         & 0.3837                              & {\color[HTML]{009901} 0.4986}       \\ \hline
Recall                                                   & {\color[HTML]{009901} 0.4769}       & \multicolumn{1}{l|}{0.4757}                        & 0.3689                              & {\color[HTML]{009901} 0.4379}       & 0.6743                              & \multicolumn{1}{l|}{{\color[HTML]{009901} 0.6793}} & 0.4316                              & {\color[HTML]{009901} 0.4770} & {\color[HTML]{009901} 0.7829}       & \multicolumn{1}{l|}{0.7138}         & 0.3847                              & {\color[HTML]{009901} 0.4639}       \\ \hline
F1-score                                                 & 0.4798                              & \multicolumn{1}{l|}{{\color[HTML]{009901} 0.4845}} & 0.3555                              & {\color[HTML]{009901} 0.4022}       & 0.6924                              & \multicolumn{1}{l|}{{\color[HTML]{009901} 0.6930}} & 0.4148                              & {\color[HTML]{009901} 0.4431} & {\color[HTML]{009901} 0.6892}       & \multicolumn{1}{l|}{0.6530}         & 0.3842                              & {\color[HTML]{009901} 0.4807}       \\ \hline
UAR                                                      & {\color[HTML]{009901} 0.6081}       & \multicolumn{1}{l|}{0.6058}                        & 0.5206                              & {\color[HTML]{009901} 0.5578}       & 0.7571                              & \multicolumn{1}{l|}{{\color[HTML]{009901} 0.7611}} & 0.5731                              & {\color[HTML]{009901} 0.5929} & {\color[HTML]{009901} 0.8077}       & \multicolumn{1}{l|}{0.7364}         & 0.5422                              & {\color[HTML]{009901} 0.6046}       \\ \hline
\end{tabular}
\end{center}
\end{table*}

\begin{figure*}[htbp]
\centering
\begin{subfigure}[b]{\textwidth}
    \centerline{
    \includegraphics[width=\textwidth]{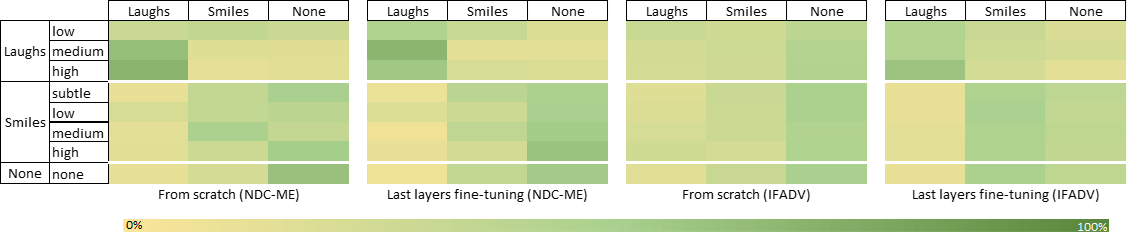}}
    \caption{Heatmaps: Audio models}
    \label{fig:Audio_heatmaps}
\end{subfigure}
\hfill
\begin{subfigure}[b]{\textwidth}
    \centering
    \includegraphics[width=\textwidth]{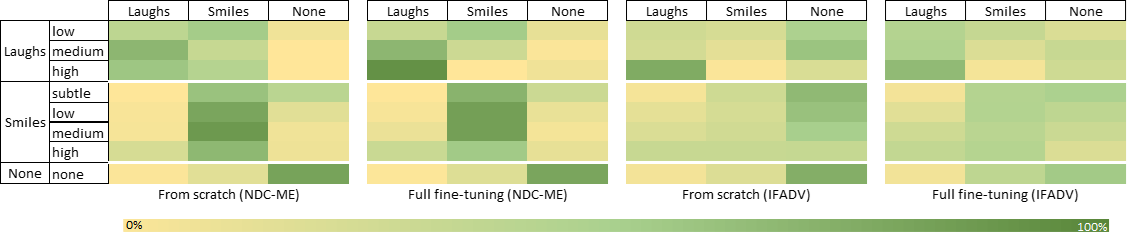}
    \caption{Heatmaps: Visual models}
    \label{fig:Video_heatmaps}
\end{subfigure}
\hfill
\begin{subfigure}[b]{\textwidth}
    \centering
    \includegraphics[width=\textwidth]{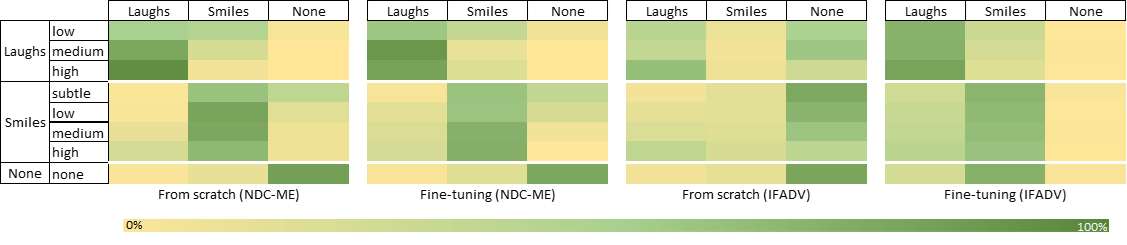}
    \caption{Heatmaps: Fusion models}
    \label{fig:Fusion_heatmaps}
\end{subfigure}
\caption{Class distribution heatmaps. Each column corresponds to the predicted class while each row shows the ground-truth label and its intensity. The colour gradient expresses the distribution in percentage per row (the sum of each row should be 100\%).}
\end{figure*}

Fig.~\ref{fig:Audio_heatmaps}, Fig.~\ref{fig:Video_heatmaps} and Fig.~\ref{fig:Fusion_heatmaps} each contains four heatmaps representing, just like confusion matrices, the way the \textbf{Laughs}, \textbf{Smiles} and \textbf{None} were classified but breaking down the classification results of the laughs and smiles into their different intensity levels. Each column shows the class predicted by the model, while each row is the ground truth intensity levels of each expression. The values are presented in percentages of the sum of the corresponding row/ground truth: for example, X\% in (row 1, column 1) represents X\% of the sum of row 1. 

In this work, we present and discuss only the results of the configurations for which we were able to draw interesting conclusions. But the readers should note that all the configurations not reported for a specific modality but that were reported for another one were indeed carried on during the experimentation phase but left out of this paper on purpose. This is for similar reasons as the fusions configuration in Section~\ref{experiment}: the results not reported here, did not bear any interesting conclusion or the performances were not good enough to even be relevant for this work (for example fine-tuning all the layers the audio modality gave very poor classification results, no interpretation of the results could be made, and so was not reported here).

Fig.~\ref{fig:Audio_heatmaps} shows heatmaps related to the audio models trained either from scratch or by fine-tuning the last layers only. The first and second heatmaps from the left are the results of the tests on the NDC-ME data. We can observe that the medium and high levels of laughs have more True Positives than the other cases for both models, and that the fine-tuned model achieves higher results for low level laughs than the model trained from scratch. The third and fourth heatmaps are the results for the same models in the first and second heatmaps (trained on NDC-ME data) but tested on IFADV data. The model trained from scratch tends to classify the samples mostly into the \textbf{None} category, while the fine-tuned model shows better performance for all laughter intensity levels.

Heatmaps in Fig.~\ref{fig:Video_heatmaps} shows the results for the visual models trained either from scratch or by fine-tuning all layers. The first and second heatmaps from the left show that both models have similar results for lower levels (subtle and low) smiles and for \textbf{None}. The model from scratch seems to confuse laughter with smiles no matter the intensity, while the fine-tuned model performs better for high level laughs. The third and fourth heatmaps contain the results for the same models mentioned before but applied on the IFADV data. The model trained from scratch shows, as expected since applied on a different dataset, decreased performances but the fine-tuned one seems to perform well on laughs. It seems to improve the performance on smiles compared to the model trained from scratch.

Fig.~\ref{fig:Fusion_heatmaps} displays the performance of the fusion of the models trained from scratch for audio and visual modalities (labelled as "From scratch") and the fusion of the fine-tuned audio and visual models mentioned above (labelled as "Fine tuning").
The first and second heatmaps show that the fusion appears to have similar results on NDC-ME data, with better a performance on high level laughter with the models trained from scratch. The third heatmap shows that the low generalisation rate of both modalities trained from scratch induces the same behaviour on their fusion model, while the fourth one presents better True Positives on both laughter and smiles detection.

\subsection{Discussion}
Firstly it is clear that not one model performed better than all the others in all categories.
But by considering overall results, we can argue that, when training and testing on the same dataset, models fusion trained from scratch performs relatively well on all classes, even better than the fine-tuned visual model which, interestingly, seems to confuse low level laughs with smiles. The fusion model seems able to keep the overall good performances of the visual modality while improving the bad ones (low level laughs notably). It is worthy to note though, that in this work, a simple fusion mechanism and training were applied. Improving these should allow to take better advantage of both modalities.
We can also note that audio laughs, when misclassified, are most often confused with smiles, especially low-level laughs. Which is an interesting point suggesting that a relationship might exist even in the audio modality.
However, this modality does not perform as well on smiles, for either evaluation datasets. It is true that the smiles true positives are quite high but so are the false positives represented by the \textbf{None} being misclassified as smiles. On an intuitive level, this makes sense. Indeed, although smiles have been shown to be audibly recognisable, smiled speech is more a change of voice than a burst of affect as is laughter, which makes it more complicated to discriminate from non-smiling speech, especially with the limited amount of data at our disposal.
The audio modality seems to perform rather well on laughter, but the smile misclassification leads to poor metrics value.
The visual models seem to perform overall better for the smiles than audio modality. This also intuitively makes sense since an obvious discriminating feature between smiles and laughs is the audio cue. Nevertheless the models also seem to perform rather well on laughs especially when fine-tuned, this is probably due to the physical movements accompanying the laughs that are less present when smiling.

Some interesting notes can also be taken concerning the fusion. First, the fusion surprisingly seems to work better when fused models were trained from scratch, than from fine-tuned ones. This fusion of models trained from scratch seems to allow the model to use the best prediction of both modalities in one system by improving the recall at the cost of a decrease in precision.
Another interesting point to note regarding fine-tuning in general is that it improves laughter classification and generalisation (when applied to IFADV data) in all cases. It also seems to improve smiles true positives score but at the expense in some cases (audio and fusion - fourth heatmap from the left) of the smiles false positives, represented by the confusion of \textbf{None} with smiles.
For the visual modality, fine-tuning seems to improve the performance of the models both for smiles and laughs detection and its generalizability most of the time which is observed on the results of the models on the IFADV data. The only slight deterioration that we can observe is that more \textbf{None} samples are confused for smiles than in the model trained from scratch.
We can deduce that fine-tuning allows a model to use the knowledge gained from prior training on speech or lip-reading data to increase its robustness to other datasets.

\begin{figure}[htbp]
    \centering
    \includegraphics[width=0.48\textwidth]{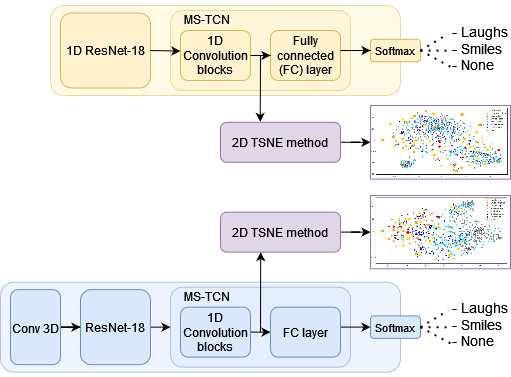}
    \caption{t-SNE dimensional reduction applied to each modality. Graphic results are display in Fig.~\ref{fig:TSNE_A} and \ref{fig:TSNE_V}.}
    \label{fig:TSNE_arch}
\end{figure}
\begin{figure*}[htbp]
\centering
\begin{subfigure}[l]{0.49\textwidth}
    \centerline{\includegraphics[width=\textwidth]{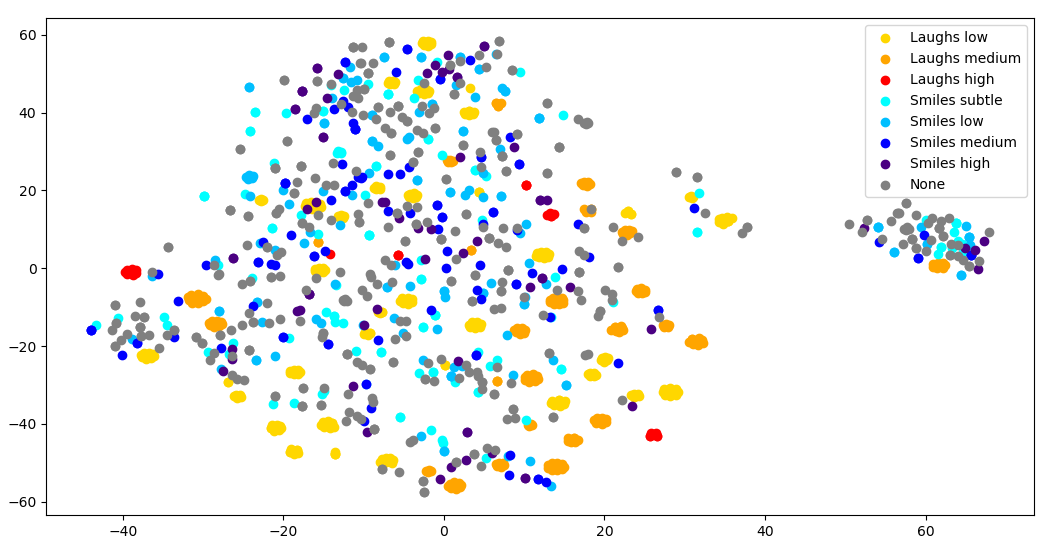}}
    \caption{Model trained from scratch.}
    \label{fig:TSNE_AS}
\end{subfigure}
\hfill
\begin{subfigure}[r]{0.49\textwidth}
    \centerline{\includegraphics[width=\textwidth]{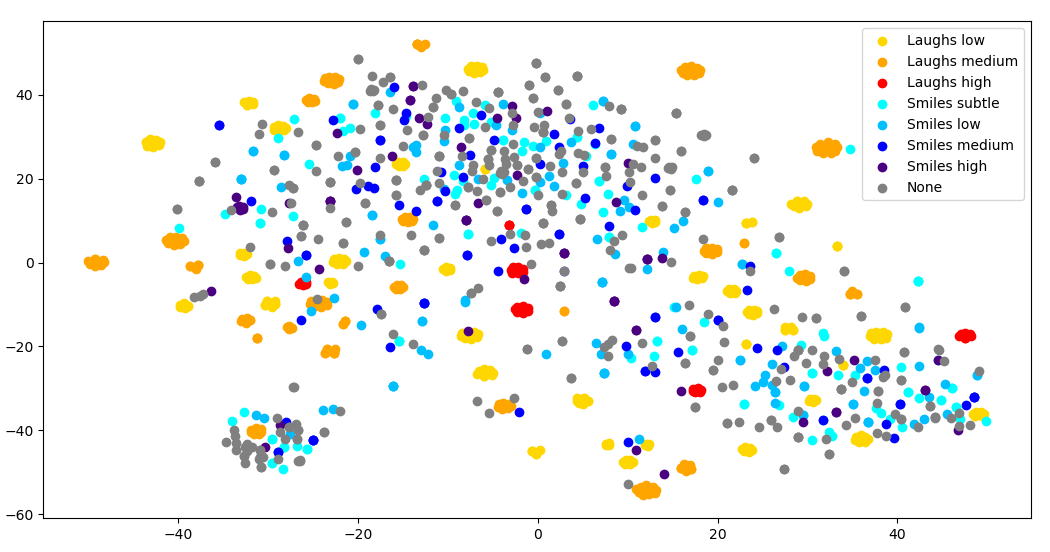}}
    \caption{\textit{Last-layers} fine-tuned model.}
    \label{fig:TSNE_AF}
\end{subfigure}
\caption{Audio models outputs using a 2D t-SNE representation.}
\label{fig:TSNE_A}
\end{figure*}
\begin{figure*}[htbp]
\centering
\begin{subfigure}[l]{0.49\textwidth}
    \centerline{\includegraphics[width=\textwidth]{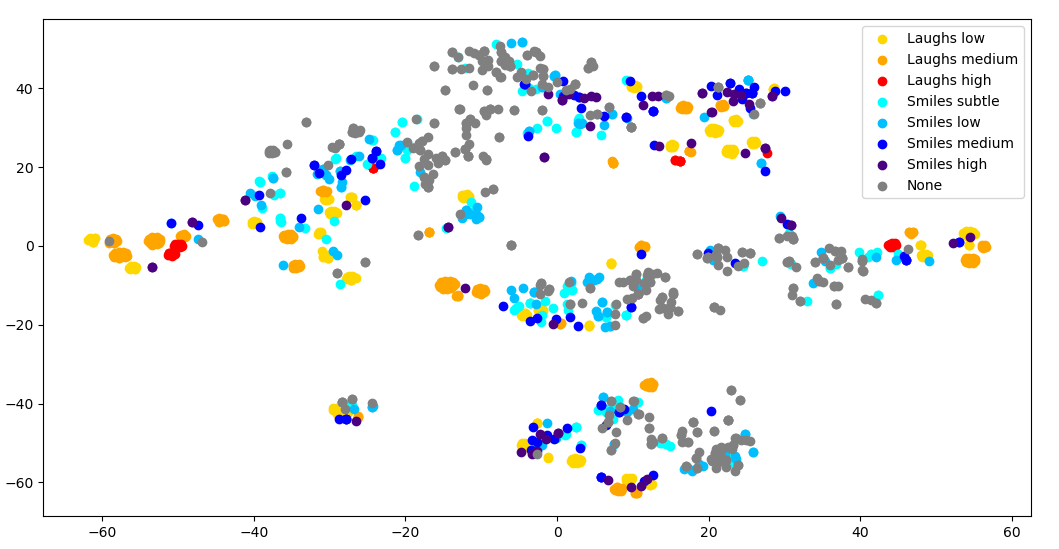}}
    \caption{Model trained from scratch.}
    \label{fig:TSNE_VS}
\end{subfigure}
\hfill
\begin{subfigure}[r]{0.49\textwidth}
    \centerline{\includegraphics[width=\textwidth]{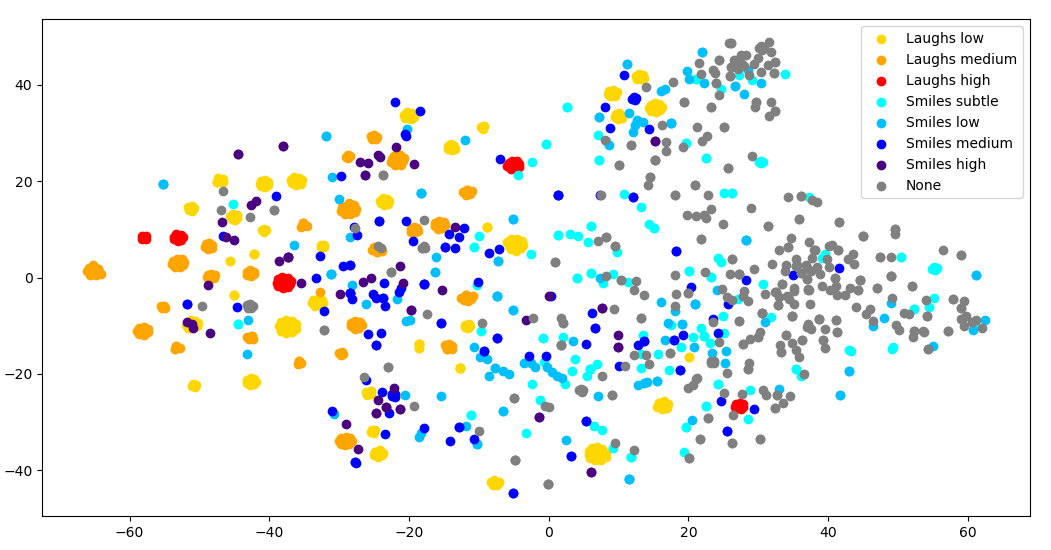}}
    \caption{\textit{Fully} fine-tuned model.}
    \label{fig:TSNE_VF}
\end{subfigure}
\caption{Visual models outputs using a 2D t-SNE representation.}
\label{fig:TSNE_V}
\end{figure*}

With the goal to get a better understanding of the models' data representation especially on the impact of fine-tuning, we present a visualisation of the ending layers of the models. For this, we extract embeddings from the output of the MS-TCN block's last layer. We then apply a t-SNE~\cite{van2008visualizing} method to reduce the embeddings dimensions to a  two-dimensional space while retaining the most relevant features. The general process is depicted in Fig.~\ref{fig:TSNE_arch} while the results on the audio modality are shown in Fig.~\ref{fig:TSNE_A} and the ones on the visual modality in Fig.~\ref{fig:TSNE_V}. For both modalities, we can see that fine-tuning allows to discriminate better the three classes. Indeed the audio laughs (shades of orange on the figure) are pushed at the extremities of the pattern, while the smiles are still rather mixed with the \textbf{None} class, which is coherent with the results presented above.
An even more interesting observation can be made on the visual data: we can clearly see the laughs being pushed at the left of the pattern, the low level laughs (yellow dots) tend to also be present in the centre of the pattern, the higher level smiles (darker blue) tend to be more mixed with the laughs and lower level smiles (lighter blue) with the \textbf{None} - all coherent with our observations made above.

An analysis of the results with respect to intensity levels shows that the system tends to learn implicit knowledge of those levels from the data. Apart from the high level laughs, the levels on the extremes seem to be more often confused by the models than the medium ones. Low level laughter are in general mostly confused as smiles and the high levels of smiles (medium and high) are mostly confused as laughs while the low levels (subtle and low) are mostly confused as being \textbf{None} (which, as we could observe in our dataset, contains a majority of neutral expressions or speech). These observations can be seen in almost all the presented results from the visual modality. This confusion by models are intuitive to us. Indeed, although they were not given any information about the intensity levels of the expressions during training, the models seem to have more difficulty with some intensities on the extreme levels than with others. In the visual modality, if we revised the current results by considering the samples classified as higher levels of smiles (medium and high) as laughs and lower level smiles (subtle and low) as \textbf{None} (thus having only 2 classes at the end instead of 3), the data would be correctly classified as laughs at an average 69.15\% with an standard deviation of 5.58\% compared to the current average rate 66.46\% with an standard deviation of 10.06\%. We assume that this is due to the nature of the expressions themselves, since the features representative of some intensities in one expression can be shared with features in another expression (high level smiles and laughs can both show pulled lip corners and raised cheeks for instance).

We can therefore mainly conclude that:
\begin{enumerate}
    \item Fine-tuning is beneficial for performance and generalization in most cases and should be considered instead of training a model from scratch.
    \item Given all the observations and analyses made regarding the intensity levels, we can safely conclude that the relationship between smiles and laughs is not as simple as a binary or single class relationship. A more complex relationship should therefore be considered when dealing with these expressions,
\end{enumerate}

Finally, the readers should note that we suspect some aspects of the dataset to have probably influenced negatively some of the results. A first aspect is that some files contain speech coming from the interlocutor of the concerned subject, overlapping the subjects laughter. More accurate detection could be achieved by removing those artefacts from the dataset. A second drawback is that some of the annotations contain subjectivity due to the limited number of annotators and this can make the annotations more sensitive to human error.

\section{Conclusion and Future Works}
\label{conclusion}

In this work, we presented a study on deep-learning based \snl classifiers discriminating laughs and smiles as two separate entities. We investigated models applied to word recognition and lip reading applications for audio and visual-based \snl classification and we fused both modalities with a simple network of fully connected layers. We showed that the fusion system achieves better overall performance than single modalities. We also highlighted the influence that fine-tuning has on the generalization of all models to different datasets. We analysed the behavior of the models with respect to the \snl intensity levels and concluded that these levels should be considered in \snl detection as they have an impact on the models trained in all modalities considered here, even though no prior knowledge was given about them to the model during training. 

We intend to investigate in future works other fusion approaches and their effect on classification, trying to improve the fusion results and focus on improving the \snl detection system's efficiency. The observations regarding the behavior of the models with respect to the intensity levels suggested that more complicated relationships between smiles and laughs should be considered rather than a simple binary one or grouping them in a single class.  We will therefore also study different grouping methods in different contexts to optimize the use of the datasets.
Finally we will integrate intensity level knowledge in the training process and analyse their effect on the knowledge acquired by the models per modality and the impact on the performances.

\section*{Ethical Impact Statement}
Several aspects should be considered when using data representing emotional states or that could be used as biometric information. In this section we attempt to draw the attention of the readers to a non-exhaustive set of aspects to take into account.

\snl are socio-culturally dependant expressions. Therefore, using them in applications without taking the context into account (user gender, socio-cultural background, interaction topic, etc.) could have unintended impacts, for example by changing the message intended or by extracting false information in a user profiling app. 

The resources available in \snl related work in general are limited due to the difficulty of collecting and annotating these expressions and their context. This makes it difficult to generalise the results obtained. Even in this work, although efforts were put in this to balance our datasets for the purposes of our experiments, improvements are still needed and even though the results suggested here provide good directions for future work, a particular attention should be put on the diversity of the subjects in the data and not only its quantity.

\snl have been considered in previous work for biometric identification purposes \cite{folorunso2020laughter, HARSHAVARDHAN2020}. It is therefore important that the field of \snl detection in general should evolve in a direction preserving the users privacy.

\bibliographystyle{IEEEtran}
\bibliography{biblio.bib}

\end{document}